\documentclass[APA,STIX2COL]{WileyNJD-v2}
\usepackage{tabularx}

\usepackage{balance}

\usepackage{boites}
\usepackage{amsmath,graphicx,color,bm}
\usepackage{framed,tikz}	
\usepackage{natbib}
\usepackage{siunitx}

\newcommand{\N}{\mathcal{N}}

\graphicspath{{figures/}}

\makeatletter
\def\NAT@aysep{,}
\makeatother

\articletype{Regular Research Paper}%

\begin{document}

\title{Probabilistic Map Matching  for Robust Inertial Navigation Aiding}

\author[1]{Xuezhi Wang*}

\author[2]{Christopher Gilliam}

\author[1]{Allison Kealy}

\author[3]{John Close}
\author[4]{Bill Moran}

\authormark{AUTHOR ONE \textsc{et al}}

\address[1]{\orgdiv{School of Science}, \orgname{RMIT University}, \orgaddress{\state{Vic 3000}, \country{Australia}}}

\address[2]{\orgdiv{School of Engineering}, \orgname{RMIT University}, \orgaddress{\state{Vic 3000}, \country{Australia}}}

\address[3]{\orgdiv{Department of Quantum Science}, \orgname{Australia National University}, \orgaddress{\state{Canberra 2601}, \country{Australia}}}

\address[4]{\orgdiv{Department of Electrical \& Electronic Engineering}, \orgname{University of Melbourne}, \orgaddress{\state{Vic 3053}, \country{Australia}}}

\corres{Xuezhi Wang,  \email{xuezhi.wang@rmit.edu.au}}


\abstract[Summary]{Robust aiding of inertial navigation systems in GNSS-denied environments is critical for the removal of accumulated navigation error caused by the drift and bias inherent in inertial sensors. One way to perform such an aiding uses matching of geophysical measurements, such as gravimetry, gravity gradiometry or magnetometry, with a known geo-referenced map. Although simple in concept, this map matching procedure is challenging: the measurements themselves are noisy; their associated spatial location is uncertain; and the measurements may match multiple points within the map (i.e. non-unique solution). In this paper, we propose a probabilistic multiple hypotheses tracker to solve the map matching problem and allow robust inertial navigation aiding. Our approach addresses the problem both locally, via probabilistic data association, and temporally by incorporating the underlying platform kinematic constraints into the tracker. The map matching output is then integrated into the navigation system using an unscented Kalman filter. Additionally, we  present a statistical measure of local map information density --- \emph{the map feature variability}  --- and use it to weight the output covariance of the proposed algorithm. The effectiveness and robustness of the proposed algorithm are demonstrated using a navigation scenario involving gravitational map matching.}



\maketitle


\section{Introduction}
\label{sec1}


In GNSS-denied (or contested) environments, platform navigation performance is dominated by the accuracy of onboard inertial sensors. Even with high end inertial sensors, which exhibit extremely low bias and drift, it is not possible to avoid the build up of navigation errors over long time frames~\citep{Titterton2004a}. Removing these accumulated navigation errors is crucial to retain the confidence of navigation accuracy~\citep{Groves2013b}. This removal, or correction, is achieved using one or more aiding sources that provide positional information, i.e. a position fix. Aiding sources can be categorised into three groups based on the technologies involved: 1) Radio-based aiding, which uses a transmitted ratio signal to obtain a position fix -- the canonical example is GNSS-based aiding. 2) Electromagnetic imaging, such as visual camera systems or Synthetic Aperture Radar (SAR) imaging, which obtain a position fix by imaging the terrain around the platform and registering the corresponding image to known landmarks. 3) Geophysics-based aiding, which obtain a position fix by measuring geophysical quantities and matching these measurements to a known geo-referenced map -- a process known as \textit{map matching}. Examples of the geophysical quantities, and their associated maps, include one or more of the elements of the gravitation vector and/or the gravity gradient tensor; the corresponding magnetic quantities; and bathymetry. In this paper, we focus on this geophysics-based aiding and present a novel map matching algorithm.

Map matching techniques are widely used in localisation and navigation scenarios where GNSS is not readily available such as underwater, urban or hostile environments. Based on how the measurements and corresponding maps are used, approaches to geophysical map matching navigation can be split into two groups: implicit map matching and explicit map matching. Implicit map matching techniques feed the geophysical measurements directly into statistical filters along with the inertial measurements. In this framework, the geo-referenced maps are used as look-up functions to compute the predicted geophysical measurements in the prediction step of the statistical filters. Due to the non-linear relationship between the estimation states and geo-physical measurements, early approaches opted to use extended Kalman filters (EKF) to perform the estimation; examples include EKFs involving gravimetry~\citep{Affleck1990}, gravity gradiometry~\citep{Jekeli2006}, or both for submarine navigation~\citep{Moryla}. A performance analysis of a gravity gradiometry EKF was presented in~\citep{lee2015performance}. More recently, to avoid the linearisation present in the EKF, unscented Kalman filters (UKF) have been proposed for gravimetry~\citep{Wu2010} and gravity gradiometry~\citep{Gao2021}. Finally, particle filters have been proposed for terrain-aided navigation using bathymetry data~\citep{teixeira2017robust}. A limitation of implicit map matching techniques however is that the statistical filters need to be re-designed when either changing the type of geophysical measurements used or incorporating new geophysical quantities. A more flexible approach is found in explicit map matching.

\begin{figure}[t]
  \centering
  \includegraphics[width=0.48\textwidth]{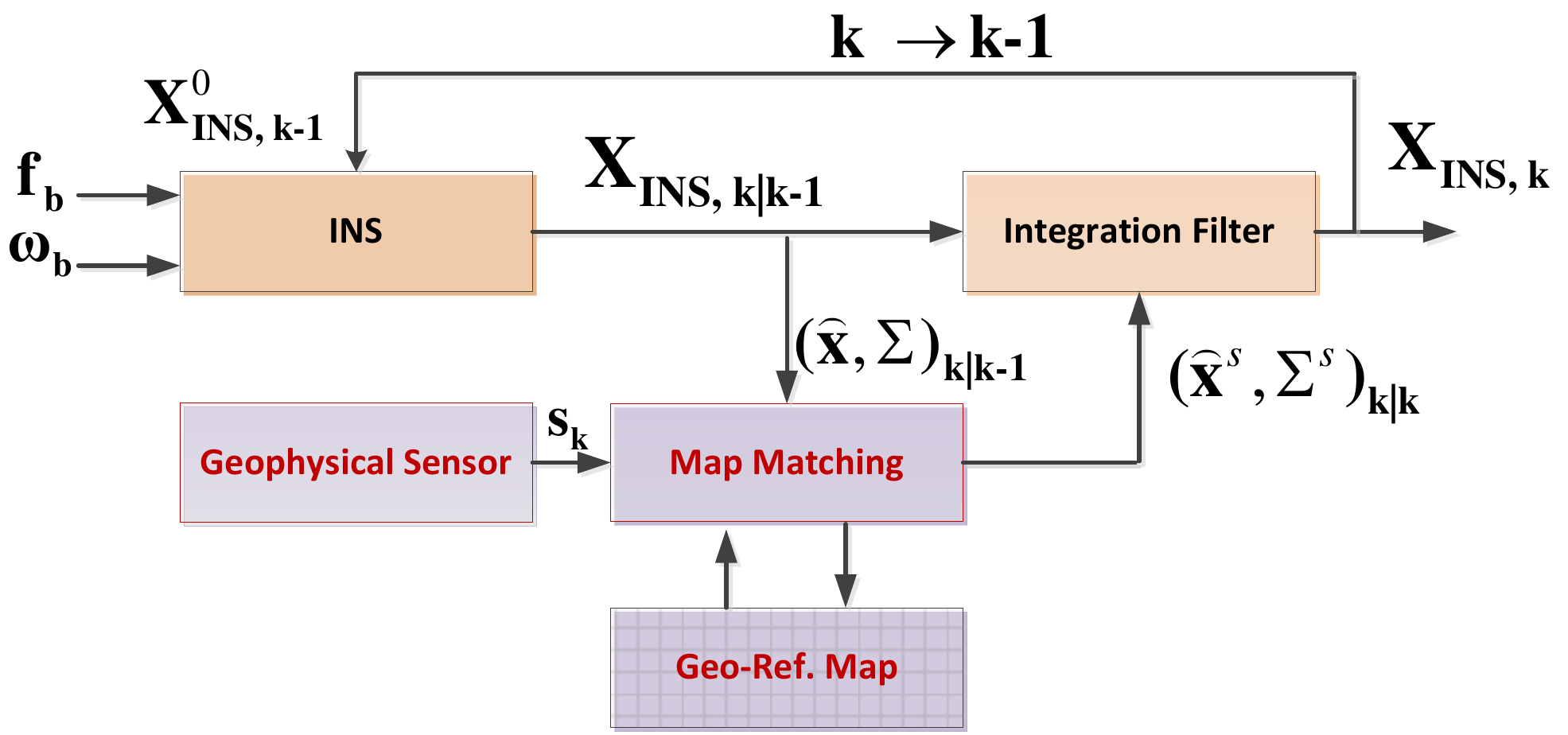}\\
  \caption{Illustration of a generic single recursion map matching aided Inertial Navigation System (INS). The map matching block takes the current kinematic state estimate and covariance of the platform, $(\hat{\bm{x}}, \bm{\Sigma})$, and augments its position component by matching the geophysical measurement $\bm{s}_k$ to a geo-reference map. The kinematic state estimate, $\hat{\bm{x}}$, is obtained by integrating the accelerator and gyroscope measurements, respectively $\bm{f}_b$ and $\bm{\omega}_b$, in the INS block. Finally, $\bm{X}_{\rm{INS}}$ denotes the full state vector from the INS and $(\hat{\bm{x}}^s,\bm{\Sigma}^s)$ represents the estimated platform position and error covariance using $\bm{s}_k$ by the map matching.}\label{fig-02}
\end{figure}

Explicit map matching techniques determine an estimate of the platform's location by directly matching the geophysical measurement to a point in the map, i.e. the matching occurs explicitly in the map space. The resulting location estimate is then integrated into the navigation system in a similar way to a loosely coupled GNSS/INS system. Figure~\ref{fig-02} shows a block diagram of a generic inertial navigation system (INS) with aiding from an explicit map matching system. The key idea is to match the geophysical measurements $\bm{s}$ to a location in the map and then use this location to improve the INS position estimate $\bm{x}$. The improved position estimate $\bm{x}^m$ is then integrated into the full INS state vector $\bm{X}_{\rm{INS}}$. Although conceptually simple, this map matching procedure is challenging for the following reasons. First, the geophysical measurements themselves are corrupted by sensor noise so the measurements will not match the map exactly. Second, the measurements may match multiple points within the map (i.e. non-unique solution). Finally, the locations where the measurements were acquired is of course uncertain. We term these challenges as the \textit{map measurement ambiguity} problem and the development of a technique to resolve this problem is the main interest of this paper.

In the literature, one approach to explicit map matching is to choose a position (or set of positions) in the map that minimise a standard error, such as mean square error or mean absolute error, between the geophysical measurement (or measurements) and the values in the map within a given region. However, as noted above, this is unlikely to yield a unique solution as multiple locations in the map may match the measurements. To solve this issue, the trajectory of the positions were constrained in~\citep{Wu2015, Wu2017} using the observation that the relative INS position change is approximately equal to relative change between the true locations. In contrast, DeGregoria~\citep{DeGregoria2010} opted to constrain the problem by performing a joint minimisation over all five of the independent elements of the gravity gradient tensor. Although straightforward, these approaches do not take account of the uncertainty in both the measurements and positional estimates, nor the structure of the map. An alternative set of approaches focus on utilising the non-uniqueness of a geophysical measurement in the map space. Specifically, a single scalar measurement belongs to an iso-contour of similar values in the geo-referenced map. Using this concept, the authors of \citep{Tuohy1996} proposed a generic map matching technique for use with two or more maps; each measurement results in a different contour and the position of the platform is determined by the intersection of these contours. Measurement uncertainty was introduced by expanding the contours to a surface envelope. Building on this work, a single map approach based on iso-contours was proposed in~\citep{Kamgar-Parsi1999}. The authors posed the problem in terms of fitting a trajectory to set of iso-contours based on initial position estimates and sensor measurements. To make the problem well-posed, a stiffness regularization term was introduce to regularise the shape of the trajectory. However, linking the kinematic constraints of the platform's motion to the regularization term is not straightforward.

In this paper, we propose a probabilistic multiple hypotheses tracking map matching (PMHT-MM) algorithm to aid the onboard INS that performs platform localisation by matching the onboard gravimetric sensor measurements with a geo-referenced data map.  The INS compensation is treated as a recursive Bayesian estimation problem. At each epoch, the prior platform location distribution is obtained from the INS computed navigation state and updated by the gravimetric signal coordinates,  estimated from map matching using a UKF. Map measurement ambiguity is addressed with  the Expectation Maximisation iterative approach,  locally using a probabilistic data association,  and temporally by considering the kinematic constraints of platform motion. Simulations using online data maps demonstrate that the proposed PMHT-MM aided INS can effectively eliminate long term INS position errors caused by inertial sensor bias and drift in the GNSS denied environment.  To the best of our knowledge, the use of probabilistic multiple hypotheses tracking algorithm for map matching is novel and is a major contribution presented in this paper.

Following the introduction, the problem formulation is given in Section~\ref{sec2}. We then present the PMHT-MM algorithm for INS aiding in Section~\ref{sec3}.  In Section~\ref{sec4}, the performance of the proposed algorithm for aiding of INS using online maps are demonstrated in a realistic navigation scenario  without GNSS. Results and discussions are presented,  followed by  conclusions in Section~\ref{sec5}.

\section{Problem formulation}
\label{sec2}

\begin{figure}[htb!]
  \centering
  \includegraphics[width=0.45\textwidth]{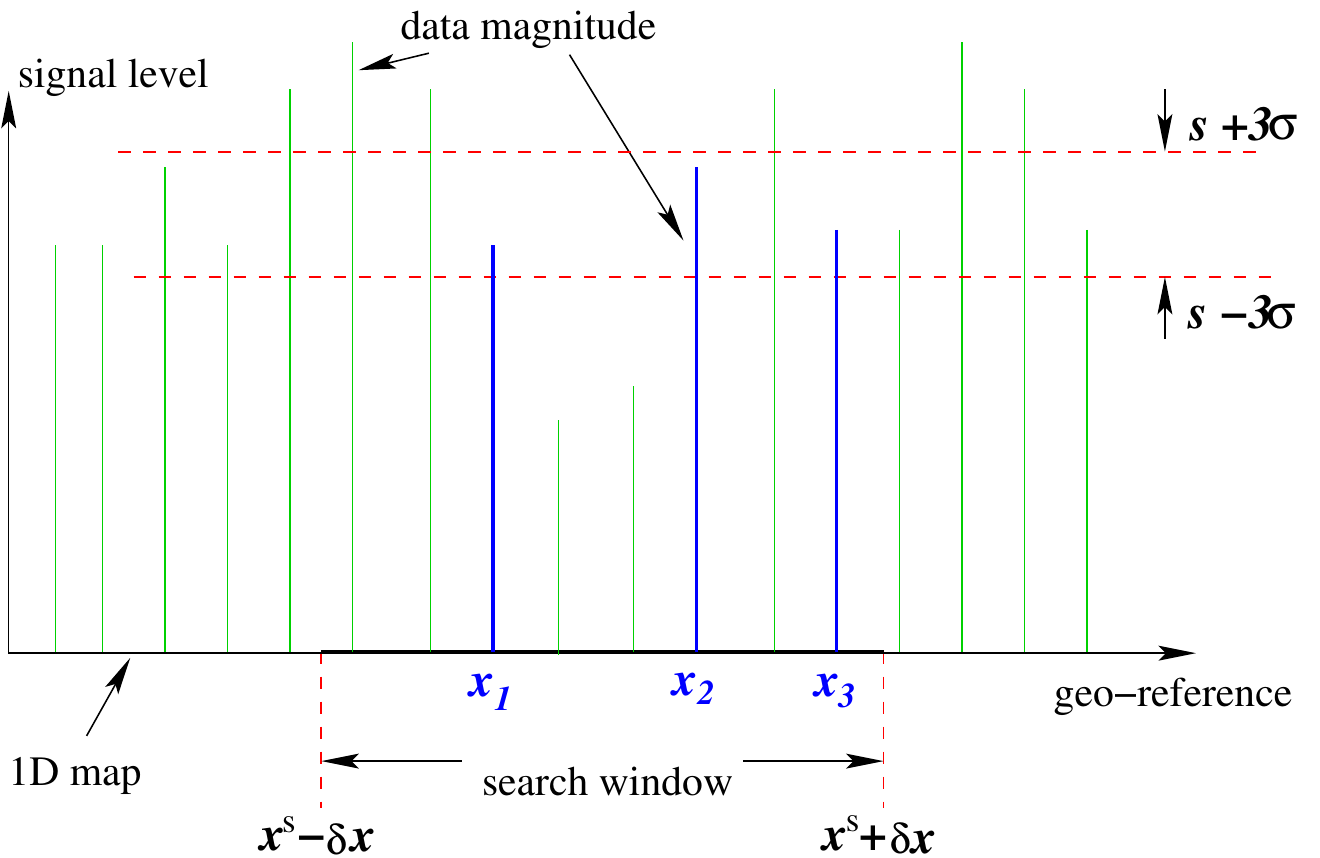}\\
  \caption{Illustration of the map lookup process by an one dimensional example. The location of the sensed signal $\bm{s}$ with standard deviation of noise $\sigma$ on the 1D map is found the search window centred at the prior of signal location $\bm{x}^s$ with uncertain off-set $\delta\bm{x}$. In this example, the collected signal location candidates on the map are $\bm{x}_1, \bm{x}_2$ and $\bm{x}_3$.}
\label{fig-54a}
\end{figure}
\begin{figure}[htb!]
  \centering
  \includegraphics[width=0.48\textwidth]{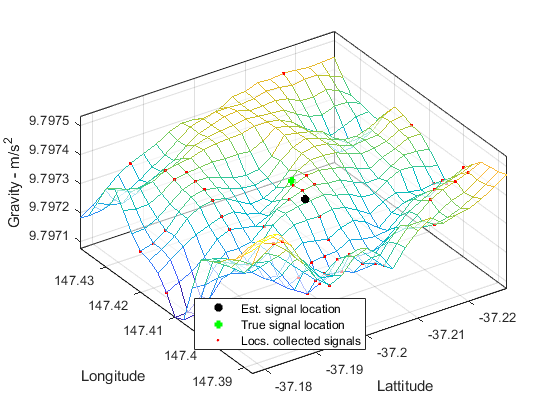}\\
  \caption{Illustration of signal location search window with the collection of signal location candidates (red dots) obtained via the map lookup function (\ref{s2-01}) with the field measurement $\bm{s} = 9.7974\, m/s^2$ and the standard deviation of measurement noise $\sigma = 0.9776\times 10^{-5}\,m/s^2$. The green dot is the true signal location and black dot signifies the PDA solution, presented next. The data map is the Global Gravity Model Plus gravity field map obtained from \citep{GGplus_online}. }\label{fig-54}
\end{figure}

Let $\bm{s}$ represent the sensed signal. This may reasonably be assumed a Gaussian distributed random variable $\bm{s} \sim \N(\bm{s}^0,\,\bm{\sigma}^2)$, where $\bm{s}^0$ is the  noiseless signal and $\bm{\sigma}$ the standard deviation of signal error.  We assume, too, that the prior distribution of platform location is Gaussian with mean and covariance being $\bm{x}^s$ and $\bm{\Sigma}^s$, respectively. The signal location from the map, denoted by $\mathcal{M}$, based on the measurement $\bm{s}$ can be expressed as
\begin{equation}\label{s2-01}
Z_m = \bm{f}(\bm{x}^s,\bm{\Sigma}^s, \bm{s}, \mathcal{M}).
\end{equation}
Equation (\ref{s2-01}) is referred to as the map lookup function. Note that the prior location distribution of the platform, together with a threshold $\gamma$,  defines an ellipsoidal area on the map centered at $\bm{x}^s$. Regardless of field measurement noise, the distribution of  the map lookup function from a single measurement $\bm{s}$ can result in more than one likely location being compatible  with  the measurement. Fig.~\ref{fig-54a} illustrates an one dimensional example of the map lookup process via (\ref{s2-01}).
We write $Z_m=\{\bm{z}_i, \, i = 1,\cdots, n\}$ for the  collection of possible  candidate locations of $\bm{s}$ from the measurement that also satisfy
\begin{equation}\label{s2-02}
 (\bm{z}_i-\bm{x}^s)(\bm{\Sigma}^s)^{-1}(\bm{z}_i-\bm{x}^s)' \leq \gamma,
\end{equation}
where $\gamma$ is a constant probability threshold.  Choice of the  value of $\gamma$  means that the ellipsoid area  contains the signal location with a certain level of confidence. In this work, we refer to such an area as a search window. It will often approximated by a rectangular area (rather than ellipsoidal); this provides significant computational efficiency with only a  minor loss of accuracy. Fig.~\ref{fig-54} shows an example of a search window on a gravity map corresponding to the down component of the gravitational vector. It spreads over an area of $5\times 5\, \mbox{km}^2$ with a collection $Z_m$ of 50 candidate locations of the measured signal with the mean of prior at the centre.

The map matching problem is to find the posterior density $p(\bm{x}|Z_m)$  of the location $\bm{x}$ of the signal $\bm{s}$ on the map $\mathcal{M}$ based on  the candidate locations, $Z_m$, and  the prior.

In this work, we propose the PMHT-MM tracker to estimate this posterior density for INS aiding. The algorithm works with a sequence of sensor measurements that corresponding to a batch of platform locations over time. Each time, the algorithm runs iteratively, using an Expectation-Maximisation technique~\citep{Dempster1977} to approximate the optimal estimator, taking into account  data correlation locally and over time through  the platform kinematic constraints.



\section{Probabilistic multiple hypotheses map matching}
\label{sec3}

In this section, we present our method to solve the map matching problem. This combines a probabilistic data association (PDA) technique from \citep{Barshalom2}, used to resolve the map measurement ambiguity issue, with a probabilistic multiple hypotheses tracker~\citep{Streit1994}, to provide a robust map matching solution in the context of INS aiding.

\subsection{Map access via probability data association}
Under the assumption that the location of a sensed signal $\bm{s}$ (e.g., obtained from the onboard INS) is of Gaussian distribution $\bm{x} \sim \N(\bm{x}^s,\,\Sigma^s)$,
the PDA solution of the map location $\bm{z}$ of signal $\bm{s}$  using the map lookup function (\ref{s2-01}) is a probabilistic combination of the set of $n$ candidate locations $Z_m = \{\bm{z}_1, \cdots,\bm{z}_n\}$. The probability weight of each candidate location $\bm{z}_i$ is proportional to the geometric distance between $\bm{z}_i$ and the window centre $\bm{x}^s$.
The probability weight can be calculated as
\begin{equation}\label{s3-01}
 w_i = \frac{p(\bm{z}_i|\bm{x}^s)}{\sum_{j=1}^n p(\bm{z}_j|\bm{x}^s)},
\end{equation}
where $p(\bm{z}_i|\bm{x}^s) \sim \N\bigl(\bm{z}_i-\bm{x}^s,\,\bm{R}_i(\sigma)\bigr)$, and $\bm{R}_i(\sigma)$ is the associated variance which is a function of the signal noise variance, or in other words, signal-to-noise ratio (SNR).
%
Thus, the PDA solution,  combining multiple locations to a single location, for the map location of sensed signal $\bm{s}$ over the area described by (\ref{s2-02}) is the following weighted mean:
\begin{equation}\label{s3-02}
\bar{\bm{z}} = \sum_{i=1}^n w_i \bm{z}_i.
\end{equation}
and the associated weighted variance:
\begin{equation}\label{s3-021}
\bar{\bm{R}} = \frac{\sum_{i = 1}^{n} w_i \bm{R}_i(\sigma)}{\sum_{j=1}^n w_j}.
\end{equation}

\subsection{PMHT-MM algorithm}
The proposed PMHT-MM algorithm is derived directly from the PMHT algorithm, originally proposed by \citep{Streit1994} for the application of multi-target tracking in clutter. We adopt this technique here for the map matching to aid an INS, where only a single target - the platform - is involved. It provides an iterative multiple hypotheses processing framework that handles measurement ambiguity locally, and system uncertainties over time under platform kinematic constraints. As pointed out in \citep{Davey2007}, it has good data association performance with a cost that is linear in time and the number of targets.

Let $\bm{x}_t$ denote the kinematic state of the platform, which involves position and velocity. Its evolution over  time is locally described by the state space model:
\begin{equation}\label{s3-03}
\bm{x}_{t+1} = \bm{F}\bm{x}_t + \bm{w}_t, \hspace{1 cm} \bm{w}_t \sim \N(0,\; \bm{Q}),
\end{equation}
and measurement model:
\begin{equation}\label{s3-04}
\bm{z}_t = \bm{H}\bm{x}_t + \bm{v}_t, \hspace{1 cm} \bm{v}_t \sim \N(0,\; \bm{R}),
\end{equation}
where $\bm{F}, \bm{H}, \bm{Q}$ and $\bm{R}$ are known matrices.

The PMHT-MM algorithm works in a batch mode  involving $T>1$ data sampling periods,  also known as \emph{scans}.  Let
$$
\bm{X} = \{ \bm{x}_1, \bm{x}_2, \cdots, \bm{x}_T \},
$$
denote the kinematic states over a batch of $T$ scans and
$$
\bm{Z} = \{ Z_1, Z_2, \cdots, Z_{T}\}
$$
be the set of measurements during the batch of scans, where $Z_t = \{\bm{z}_1, \bm{z}_2, \cdots, \bm{z}_{n(t)}\}$ signifies the set of $n(t)$ measurements collected at scan $t$.

The PMHT-MM seeks to maximise the posterior probability density function $p(\bm{X}|\bm{Z})$ by performing the following expectation-maximisation (EM) iteration
\begin{equation}\label{s3-05}
\hat{\bm{X}}^{(i+1)} = \arg \max_{\bm{X}} \Phi(\bm{X})
\end{equation}
where
\begin{equation} \label{s3-06}
\Phi(\bm{X}) = \sum_\Theta p(\Theta|Z,\bm{X}^{(i)}) \log p(\bm{Z}, \Theta|\bm{X})
\end{equation}
and  $\Theta = \{k_r(t) \}$ represents a set of
measurement-to-platform association events, e.g., $k_r(t)$ is the
event that the $r$th measurement $\bm{z}_r(t)$ at scan $t$  originates from the platform. In the EM terminology, $\Theta$ is the \emph{latent variable}, $\bm{Z}$ is called the \emph{incomplete data} and $(\bm{Z}, \Theta)$ is referred to as the \emph{complete data}. In the context of map matching, $\bm{Z}$ is the batch of $T$ subsets of platform location candidates corresponding to the field measurements observed in the $T$ scans,  and $\Theta$ represents the way of selecting the locations associated with these field measurements from $\bm{Z}$.


At  time $k = t+T$, the prior kinematic state of platform $\hat{\bm{X}}_k$, as expressed in (\ref{kstate}) for $i=0$, is obtained from the navigation state of the INS, and the set of signal candidate locations $\bm{Z}_k$ which are obtained via the map lookup function (\ref{s2-01}) based on the set of gravimeter sensed signals $\{\bm{s}_{k-T},\cdots,\bm{s}_k\}$. The PMHT-MM then runs the following two steps iteratively. At the $i$th iteration:
\begin{enumerate}
\item[Step 1:] Calculate the PDA solution of map locations $\{\bar{\bm{z}}_1, \cdots, \bar{\bm{z}}_T\}$ and their associated variances $\{\bar{\bm{R}}_1,\cdots,\bar{\bm{R}}_T\}$ for the set of sensed signals  $\{\bm{s}_{k-T},\cdots,\bm{s}_k\}$ based on knowledge of $\hat{\bm{X}}^{(i)}_k$ via (\ref{s3-01}),(\ref{s3-02}) and (\ref{s3-021}). Note that the likelihood of the $j$th candidate location $p(\bm{z}_j|\bm{x}^s)$ in (\ref{s3-01}) is replaced by
    \begin{equation}\label{s32-1}
    p(\bm{z}_j|\bm{x}^s) =  \N(\bm{z}_j; \bm{H}\hat{\bm{x}}^{(i)}_{t|t-1}, \bar{\bm{R}}_t), \hspace{0.5 cm} t = 1,\cdots,T.
    \end{equation}
    where $\hat{\bm{x}}^{(i)}_{t|t-1}$ is the $i$th iteration PMHT-MM predicted kinematic state of the platform at time $k-t$.
\item[Step 2:] State update and smoothing.
The prior state estimates and the associated covariance matrices
   \begin{equation}\label{kstate}
   \bm{x}^{(i)}_{1}, \bm{x}^{(i)}_{2}, \cdots, \bm{x}^{(i)}_{T}, \hspace{0.5 cm} \mbox{and} \hspace{0.5 cm}
  \bm{\Sigma}^{(i)}_{1}, \bm{\Sigma}^{(i)}_{2}, \cdots, \bm{\Sigma}^{(i)}_{T},
  \end{equation}
  are updated via a fixed lag Kalman smoother in a forward (update) and backward (smoothing)
  recursion using the set of measurements obtained from Step~1.
  \begin{itemize}
  \item Forward process
  $$
  \hat{\bm{y}}_{1|1} = \bm{x}^{(i)}_{1}, \hspace{0.5 cm} \mbox{and} \hspace{0.5
  cm}\bm{P}_{1|1} = \bm{\Sigma}^{(i)}_{1}
  $$
  For $t = 1, \cdots, T-1$, we have (standard Kalman filtering)
  \begin{eqnarray}
  \bm{P}_{t+1|t} &=& \bm{F}\bm{P}_{t|t}\bm{F}^{'} + \bm{Q} \nonumber \\
  \bm{K}_{t+1} &=& \bm{P}_{t+1|t}\bm{H}^{'}\bigl( \bm{H}
  \bm{P}_{t+1|t}\bm{H}^{'} + \bar{\bm{R}}^{(i+1)}_{t+1} \bigr)^{-1}
  \nonumber \\
  \bm{P}_{t+1|t+1} &=& \bm{P}_{t+1|t}- \bm{K}_{t+1}\bm{H}\bm{P}_{t+1|t} \nonumber
  \\
  \hat{\bm{y}}_{t+1|t+1} &=& \bm{F}\hat{\bm{y}}_{t|t} + \bm{K}_{t+1}
  \bigl(\bar{\bm{z}}^{(i+1)}_{t+1} - \bm{H}\bm{F}\hat{\bm{y}}_{t|t}\bigr)
  \label{forward}
  \end{eqnarray}
 \item Backward process:
  $$
  \bm{x}^{(i+1)}_{T} = \hat{\bm{y}}_{T|T}
  $$
  For $t = T-1, \cdots, 2,1,$ (smoothing)
  \begin{equation}
  \bm{x}^{(i+1)}_{t} = \hat{\bm{y}}_{t|t} +
  \bm{P}_{t|t}\bm{F}^{'}\bm{P}_{t+1|t}^{-1} \bigl(\bm{x}^{(i+1)}_{t+1} -
  \bm{F}\hat{\bm{y}}_{t|t} \bigr) \label{backward}
  \end{equation}
  \end{itemize}
\end{enumerate}
The iteration may be stopped if the criterion $\| \bm{X}^{(i+1)} -\bm{X}^{(i)}\| \leq \varepsilon $ is met, or after a fixed number of iterations. In this work, we chose the number of iterations as 15 in all simulations as, in our context,  almost no error difference between two consecutive iterations is observed after 15 iterations.

\subsection{Map matching aiding}
The PMHT-MM algorithm is designed to work locally in  coordinates consistent with the INS and map geo-reference, so that it deals with the (noisy)  linear kinematics using standard Kalman filters. In this work, the kinematic components of the INS navigation state are taken as priors and an estimate is made of the current platform kinematic state based on a batch of sensed signals taken from a gravimetric sensor independent of INS. As illustrated in Fig.~\ref{fig-02}, the posterior estimate $(\hat{\bm{x}},\, \bm{\Sigma})$ of the platform is integrated into the INS via a loosely coupled unscented Kalman filter (UKF). Interested readers may refer to \citep{Titterton2004a} and \citep{Crassidis2006} for more information on the strapdown INS with UKF integration. We highlight several points below specifically regarding PMHT-MM aiding integration.
\begin{itemize}
\item Platform kinematic behaviour may be quite complex, but locally, within the batch length $T$,  can be approximated by a linear system.  A trade-off between aiding robustness and allowable platform maneuver capability is achieved  by choosing a suitable  batch length.
\item Two alternative approaches can be used to implement the PMHT-MM algorithm:
\begin{description}
\item[Standard:] an update occurs after every batch time duration ``$T\times \Delta t$'',  for example, $ T = 30, \Delta t = 10$, then the aiding interval will be 300s, where $\Delta t$ is the gravimetric sensor sampling interval.
\item[Retrodiction:] update occurs after every ``$T\times \Delta t $'' with all navigation state components involved in the batch processing by retrodiction. This is equivalent to have an aiding interval $\Delta t  = 10s$.
\end{description}
Our simulation suggests that the estimated platform trajectory is more smoother by using retrodiction, though more computational resources are required.
\item In view of the fact that the  data variability (variation of features)   of a map varies  from place to place, it is desirable to define a measure to describe that variability and to find a way to  take this into account in the filter for map matching aiding. In this work, such a measure, called \emph{map feature variability}, is defined. It is denoted by $\mathcal{C}_i$, where $i$ indicates the pixel around which the variability is quantified. The map feature variability at the $i$th pixel (location $\bm{x}^{s}_i$) within a fixed search window template centred at $i$ is
\begin{equation}\label{CV}
\mathcal{C}_i = \frac{1}{n}\sum_{j}^n(\bm{x}^{\bm{s}}_i - \bm{x}_j)^2, \hspace{0.5 cm} \forall \, \bm{x}_j \in \mbox{search window},\, \bm{x}_j \neq \,\bm{x}^{s}_i,
\end{equation}
where $n$ is the number of points in the search window. The map feature variability for a given map location provides a confidence measure on the local map accuracy level. In practice, this quantity is normalised over a fixed number of samples. If it is too small,  it might be more effective to stop the aiding as, in these circumstances,  it contributes little and might actually impair the performance of the INS  tracker. In the aiding process, the covariance of the estimated location by PMHT-MM algorithm is weighted by the map feature variability.

Fig.~\ref{fig-60} shows an example of the map feature variability sequence computed along the platform travel path in the simulation scenario presented next. The magnitude of the map feature variability reflects the level of data variation on the map as indicated by the vertical gravity disturbance measurement sequence curve shown in the top figure along the platform trajectory taken by a noiseless sensor.
\begin{figure}[hbt!]
  \centering
  \includegraphics[width=0.48\textwidth]{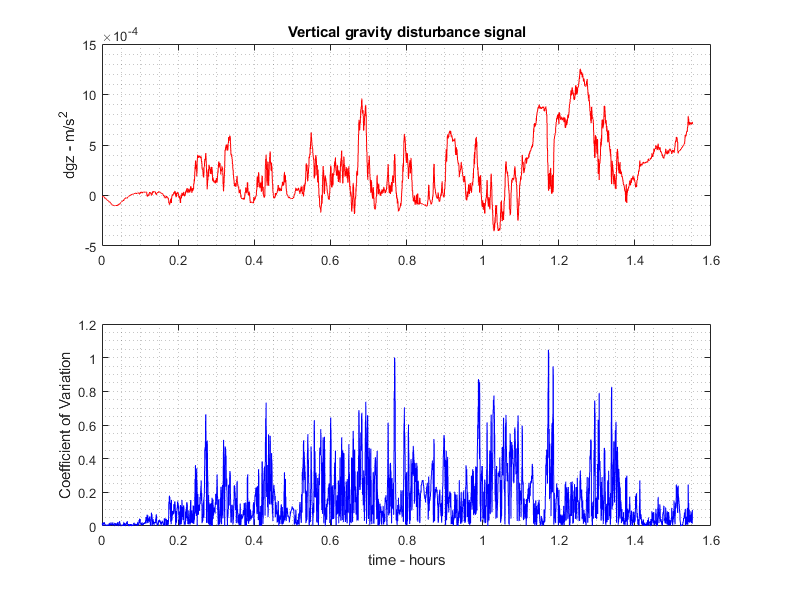}
  \caption{Noiseless vertical gravity disturbance measurements and the map feature variability of the map along the vehicle travel trajectory. Note that in practice, the vertical gravity disturbance measurement is the vertical part of the deflections from the normal gravity. In this work, we assume that the reading error of deflection angle is determined by the onboard gyroscope accuracy and is contained in the field sensor model. }\label{fig-60}
\end{figure}
\end{itemize}

\section{Experiment and results}
\label{sec4}

In this section, the proposed PMHT-MM aiding method is tested in a scenario of inertial navigation with aiding only from map matching using gravimetric sensor measurements with associated data maps. The maps used in the experiment are the ultra-high resolution, non-parametric gravity maps, known as  GGMplus~\citep{Hirt2013}.  We use two maps from GGplus:  a vertical gravity field data map  and a vertical gravity disturbance map, both   obtained online~\citep{GGplus_online}. 

\subsection{Geophysical Data}
To exemplify our algorithm, we use the ultra-high resolution, non-parametric, gravity maps presented in \citep{Hirt2013,Hirt2014}. These maps, known as Global Gravity Model Plus (GGMplus), achieve a spatial resolution of $\sim250$~m and covers all land and near-coastal areas for the Earth between $\pm60^{\circ}$ latitude. This ultra-high resolution map is obtained by fusing the following three elements:
\begin{enumerate}
    \item GOCE/GRACE satellite gravity (spatial scales of $\sim$10,000~km  down to $\sim$100~km).
    \item Global geopotential model EGM2008 (spatial scales of $\sim$100~km to $\sim$10~km).
    \item Topographic gravity due to terrain (spatial scales of $\sim$10~km to $\sim$250~m).
\end{enumerate}
Note that the topographic gravity is obtained assuming a mass density of $2670$~$\text{kg}\,\text{m}^{-3}$. For our simulations, we use gravitational acceleration in the down and the gravity disturbance (the radial derivatives of the disturbing gravity potential). These maps can be accessed at~\citep{GGplus_online}.

\subsection{Simulation scenario}

\begin{figure}[hbt!]
  \centering
  \includegraphics[width=0.48\textwidth]{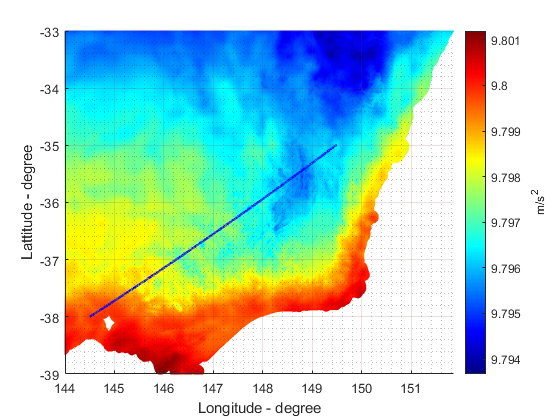}\\
  \caption{The ultrahigh resolution data map of gravity field
 obtained from \citep{GGplus_online}. The platform travel trajectory (blue line) is superimposed on the map.}\label{fig-51}
\end{figure}

The simulation scenario is a constant velocity vehicle traveling along the surface of the earth at a fixed height of 100 m from $[-38^\circ, 144.5^\circ]$ to $[-35^\circ, 150^\circ]$ (i.e., from Melbourne area to Sydney area) and at a ground speed of 22 m/s. The entire journey takes more than 3.6 hours and the PMHT-MM tracker is the only form of aiding to the onboard INS. Fig.~\ref{fig-51} shows the  vehicle travel trajectory and the geo-referenced data map used for the test.
The onboard INS computes the navigation state consisting of the platform geographical coordinates, navigation frame velocity, attitude and associated accelerometer and gyroscope biases at a sampling rate of $1$ Hz, corresponding to a standard INS implementation as described in \citep{Titterton2004a,Groves2013b}. We assume that a low noise gravimeter is onboard the vehicle to take gravity field measurements at an interval of every $\Delta t = 10$ seconds. As mentioned earlier, a UKF is used to incorporate the PMHT-MM output to update navigation state of the INS at an interval of $T \times \Delta t$. For comparison, we choose the batch lengths of PMHT-MM $T = 15$ and $T=30$, respectively.

The primary simulation objectives is to compare the performance of the INS equipped high end inertial sensors with that of the INS aided by PMHT-MM to see what level of bias correction can be acquired after the PMHT-MM aiding. The performance is measured using the metric of root-mean-squared (RMS) position error on the vehicle trajectories. In addition, track divergence rate in multiple  Monte Carlo  runs is counted as an indication of the robustness of the PMHT-MM. A track in a single run is deemed to be divergent if the RMS position error becomes increasingly large over time without apparent bound.  PMHT-MM tracker divergence is caused by a large uncertainty in a search window center estimation with a short batch length $T$; this  results in repeated exclusion  of true signal locations in the search windows.

100 Monte Carlo runs are carried out for the INS with each of the following two sets of inertial sensors:
\begin{enumerate}
    \item Precision grade accelerometer and gyroscope (PCAG);
    \item Quantum grade accelerometer and precision grade gyroscope (QAPCG);
\end{enumerate}
The noise characterisation of the inertial sensors is given in Table~\ref{table01}. 
\begin{table}[ht]
\centering
\caption{Bias and noise ranges of inertial sensors in the simulation according to \citep{2005:Christopher_Jekeli}.} \label{table01}
\resizebox{\columnwidth}{!}{
\begin{tabular}{|l|l|c|c|}
  \hline
  Sensor Grade & Sensor Type & Bias $b$ & White Noise $\sigma$ \\ \hline
 Precision (PC)  & Accel. horiz. & $2\times 10^{-6} m/s^2$ & $8\times 10^{-5} m/s^2/\sqrt{Hz}$ \\\cline{2-4}
   & Accel. Vert. & $2.5\times 10^{-8}m/s^2$ & $1.6\times 10^{-6} m/s^2/\sqrt{Hz}$ \\\cline{2-4}
   & Gyro. horiz.  & $2\times 10^{-5} deg/h$ & $1\times 10^{-3} deg/h/\sqrt{Hz}$ \\\cline{2-4}
   & Gyro. vert. & $1\times 10^{-3} deg/h$ & $3\times 10^{-2} deg/h/\sqrt{Hz}$ \\ \hline
  Quantum (QS) & Accel. & $1\times 10^{-8} m/s^2$ & $3\times 10^{-8} m/s^2/\sqrt{Hz}$ \\ \cline{2-4}
   & Gyro. & $1\times 10^{-5} deg/h$ & $1.2 \times 10^{-4} deg/h/\sqrt{Hz}$ \\
  \hline
\end{tabular}}
\end{table}

In addition, 100 Monte Carlo runs are carried out for each of the PMHT-MM aided INS cases:
\begin{enumerate}
\item Batch length $T = 15$ and the standard deviation of gravimetric sensor noise is $\sigma = 10^{-5}\,m/s^2$  or $\sigma = 2\times 10^{-4}\, m/s^2$;
\item Batch length $T = 30$ and the standard deviation of gravimetric sensor noise is $\sigma = 10^{-5}\,m/s^2$  or $\sigma = 2\times 10^{-4}\, m/s^2$.
\end{enumerate}
where precision grade inertial sensors (PCAG) are used. Both low and high noise levels are chosen for the gravimetric sensor in the simulation. The sensor of low noise level represents a best possible high end field sensor. On the other hand, the value of high noise level is chosen such that below which the PMHT-MM algorithm with the underlying map will work robustly without divergence.

\begin{figure}[hbt!]
  \centering
  \includegraphics[width=0.48\textwidth]{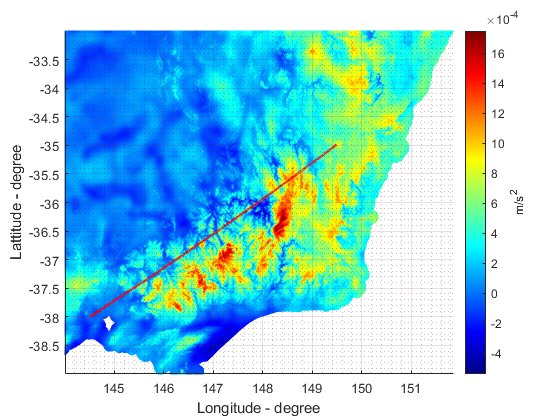}\\
  \caption{Ultrahigh resolution data map of gravity disturbance (unit: $m/s^2$) downloaded from \citep{GGplus_online}. The platform trajectory (red line) is superimposed on the map.}\label{fig-57}
\end{figure}

In the map matching, at each map location of the measured signal $\bm{s}$ ``predicted'' by the onboard INS, a set of up to 20 locations (of data values closest to $\bm{s}$)  are collected via the map lookup function (\ref{s2-01}). The average size of search windows is about $5\times 5\, \mbox{km}^2$ for using the gravity field map shown in Fig.~\ref{fig-51} and this number is slightly larger for using the gravity disturbance map shown in Fig.~\ref{fig-57}.

The above mentioned simulations are also repeated with the map matching using the gravity disturbance map, which has a larger grid cell (thus lower resolution) than that of the gravity field map as shown in Fig.~\ref{fig-57}. In that case, the gravity disturbance measurement at each epoch is obtained by processing of the measurement of onboard gravimetric sensor. Two sensor noise levels are considered: the standard deviations of the sensor noise are $10^{-6} m/s^2$ and $4\times 10^{-5} m/s^2$, respectively.

\subsection{Results and discussions}

\subsubsection{Overview simulation results}

\begin{figure}[hbt!]
  \centering
  \includegraphics[width=0.48\textwidth]{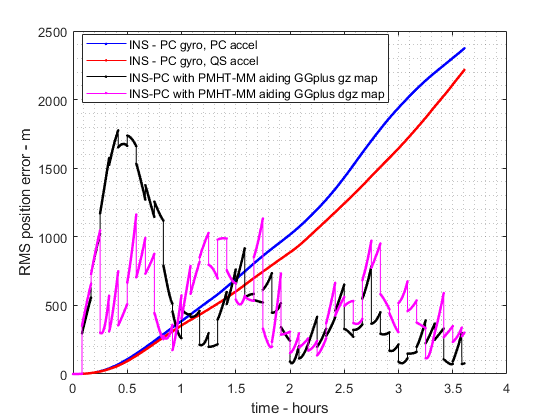}
  \caption{Comparison of RMS position errors of the INS 1) with PCAG; 2) with QAPCG; 3) with PCAG and PMHT-MM aiding using the vertical gravity field map (black curve); 4) with PCAG and PMHT-MM aiding using the vertical gravity disturbance map (pink curve).}\label{fig-52}
\end{figure}
Before going to detail, we present a summary of the simulation results in Fig.~\ref{fig-52}. The figure compares the RMS position errors of computed vehicle trajectories by the INS equipped with each of the two inertial sensor suites PCAG and QAPCG, respectively. Along them we plot the RMS position errors of the INS navigated trajectories with PCAG aided by the PMHT-MM algorithm with a batch length of $T=30$  using the two GGplus maps, respectively. Each of the results are averaged over 100 runs. In the  navigation experiment aided by the PMHT-MM a PCAG inertial sensor suite is used,  and the aiding from PMHT-MM is integrated by an UKF whose predicted state is purely based on INS.

Observations are made from Fig.~\ref{fig-52} as follows. \vspace{-0.3cm}
\begin{itemize}
\item In the INS without aiding scenario, the RMS position error caused by the accumulative bias and drift grow over time unbounded, even with the high end inertial sensors.
\item Use of the quantum grade accelerometers, which have extremely low bias and drift,  results in a  reduction of the RMS position error  by a little over $6\%$ after 3.5 hours compared  with the PCAG. Nevertheless, the position error drift is still unbounded.
\item With aiding from the proposed PMHT-MM algorithm using field measurements and data map, the position drift accumulated from the INS over time is bounded and a stable RMS position error performance can be maintained.
\item In this simulated example, high end inertial sensors are used for the onboard INS, the benefit of aiding by PMHT-MM appears after 2 hours.
\item The sampling period of the field sensors is 10 s and the aiding period of the PMHT-MM is $10\times 30 = 300 s$. On the other hand, the sampling period of the INS is 1 s. This is a partial reason that the RMS error curves of the aided INS appear a little jagged though averaged from 100 runs each. In addition, it is worth mentioning that the error bound level of the PMHT-MM aiding depends on the field sensor precision and map resolution.
\end{itemize}

\subsubsection{Result with the gravity field map}

\begin{figure}[hbt!]
\centering
\includegraphics[width = 0.48\textwidth]{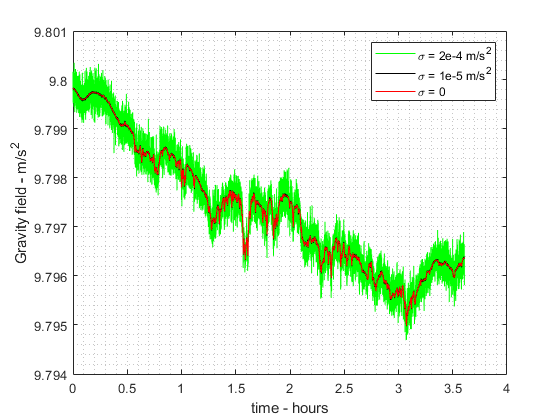}
\caption{Gravity field measurements along the vehicle trajectory with noise $\sigma = 0, 10^{-5}$ and $2\times 10^{-4}\,m/s^2$, respectively.} \label{fig-56a}
\end{figure}

\begin{figure}[hbt!]
  \centering
  \includegraphics[width=0.48\textwidth]{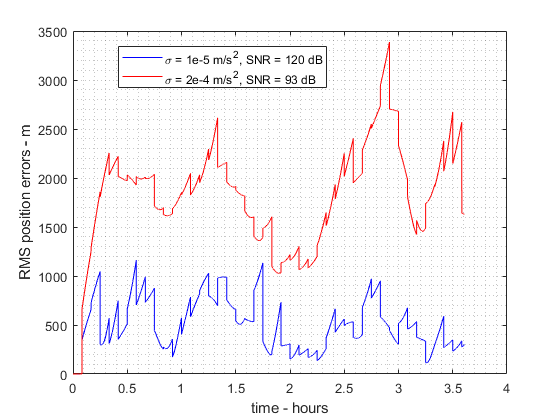}
  \caption{Comparison of RMS position error of the INS with PMHT-MM aiding for $T=30$ at measurement noise levels $\sigma = 10^{-5}\, m/s^2$ (SNR = 120 dB) and $2\times 10^{-4}\,m/s^2$ (SNR = 93 dB), respectively. }\label{fig-56}
\end{figure}

\begin{figure}[hbt!]
  \centering
  \includegraphics[width=0.48\textwidth]{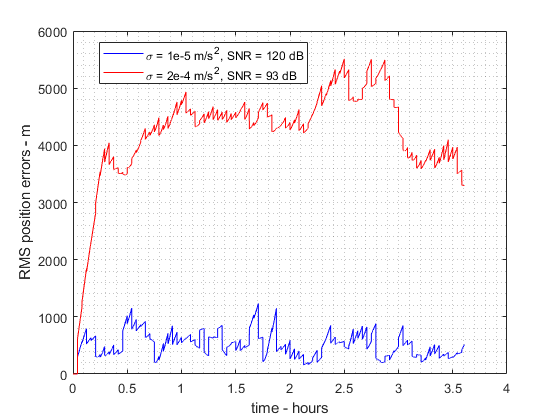}
  \caption{RMS position error of the INS with PMHT-MM aiding for $T=15$ at measurement noise levels $\sigma = 10^{-5}\, m/s^2$ (SNR = 120 dB) and $2\times 10^{-4}\,m/s^2$ (SNR = 93 dB), respectively. }\label{fig-56b}
\end{figure}

The gravimetric sensor measurement sequences along the platform travel trajectory with the standard deviations of noise $\sigma = 0, 10^{-5}$ and $2\times 10^{-4}\,m/s^2$ are shown in Fig.~\ref{fig-56a}.
The mean position error and track divergence rate under three levels of measurement noise, and averaged over 100 runs,  are shown in Table~\ref{table4}.
\begin{table}[hbt!]
\centering
\caption{Mean error and diverge rate of the INS with PMHT-MM aid using the GGplus gravity field map.  }\label{table4}
\resizebox{\columnwidth}{!}{
\begin{tabular}{|c|c|c|c|}
  \hline
  Batch Length & Mean position error & $\sigma$ ($m/s^2$)\& SNR & Diverge Rate \\ \hline \hline
  T=30 & 507 m& $\sigma =10^{-5}$, SNR = 120 dB & 0\% \\ \hline
   T=15 & 510 m & $\sigma =10^{-5}$, SNR = 120 dB & 0\% \\ \hline
  T=30  & 1820 m &$\sigma = 2\times 10^{-4}$, SNR = 93 dB & 6\% \\ \hline
  T=15  & 4199 m &$\sigma = 2\times 10^{-4}$, SNR = 93 dB & 22\% \\ \hline
  \hline
\end{tabular}}
\end{table}
The results shown in Table~\ref{table4} suggest that the position aiding output estimated by the PMHT-MM  from gravimeter measurements matched against the GGplus map yields an average position  error in excess of  500 m at the measurement noise level $\sigma = 10^{-5}\,m/s^2$ (SNR = 120 dB). This position error grows rapidly as the measurement noise level increases. Correspondingly,  the tracker divergence rate also increases. The RMS position error comparison for $T=30$ shown in Fig.~\ref{fig-56} confirms this observation.
A similar situation for $T=15$ is shown in Fig.~\ref{fig-56b}, though the error differences between the two levels of sensor noise  become even larger.
We see from Table~\ref{table4} that with low sensor noise ($\sigma = 10^{-5}\, m/s^2$)  map matching is robust with zero divergence rate for both $T=15$ and $T=30$ and RMS error performance is dependent on the quality of data map.

\subsubsection{Result with gravity disturbance map}

In the case of PMHT-MM aiding using the gravity disturbance map, the mean position error and track divergence rate under three levels of measurement noise, and averaged over 100 runs,  are shown in Table~\ref{table6}.
\begin{table}[hbt!]
\centering
\caption{Mean error and divergence rate with the GGplus gravity disturbance map. }\label{table6}
\resizebox{\columnwidth}{!}{
\begin{tabular}{|c|c|c|c|}
  \hline
  Batch Length & Mean position error & $\sigma\, (m/s^2)$ \& SNR & Diverge Rate \\ \hline \hline
  T=30 & 544 m& 1e-6, 51 dB  & 0\% \\ \hline
  T=15 & 868 m& 1e-6, 51 dB  & 13\% \\ \hline
  T=30 & 1056 m  & 4e-5, 20 dB & 13\% \\ \hline
  T=15 & 2673 m  & 4e-5, 20 dB & 69\% \\ \hline
  \hline
\end{tabular}}
\end{table}

\begin{figure}[hbt!]
  \centering
  \includegraphics[width=0.48\textwidth]{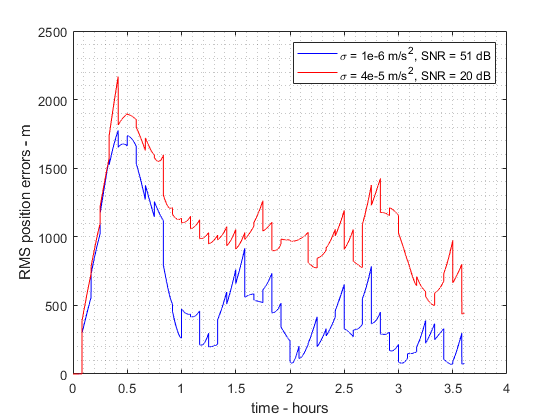}\\
  \caption{RMS position errors of the INS with PMHT-MM aiding at every 300 seconds using the GGplus vertical gravity disturbance map shown in Fig.~, where the batch length of PMHT-MM is T = 30.}\label{fig-61}
\end{figure}

\begin{figure}[hbt!]
  \centering
  \includegraphics[width=0.48\textwidth]{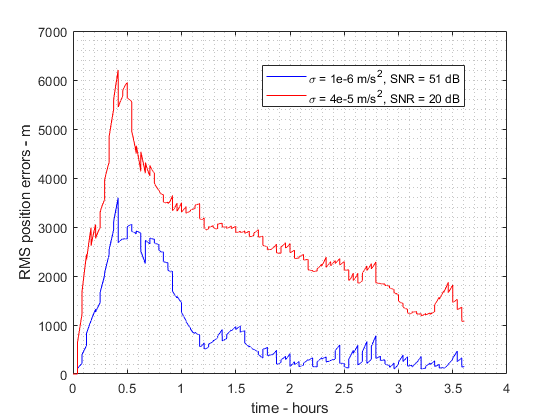}\\
  \caption{RMS position errors of the INS with PMHT-MM aiding at every 300 seconds using the GGplus vertical gravity disturbance map shown in Fig.~, where the batch length of PMHT-MM is T = 15.}\label{fig-61a}
\end{figure}

The RMS position error performances of the PMHT-MM aided INS along the platform  trajectory are shown in Fig.~\ref{fig-61} for $T= 30$ and Fig.~\ref{fig-61a} for $T = 15$, respectively.
These results obtained with the GGplus gravity disturbance map show no significant difference compared with those using the GGplus vertical gravity field map shown in Fig.~\ref{fig-56} and Fig.~\ref{fig-56b}. In addition,  map-dependent PMHT-MM aiding accuracy is clearly observed from the RMS position error of vehicle first hour trip; this is overwhelmingly large because of a small map feature variability in that area evidenced by the map feature variability along the trajectory shown in the bottom plot of Fig.~\ref{fig-60}.

\begin{figure}[hbt!]
  \centering
  \includegraphics[width=0.48\textwidth]{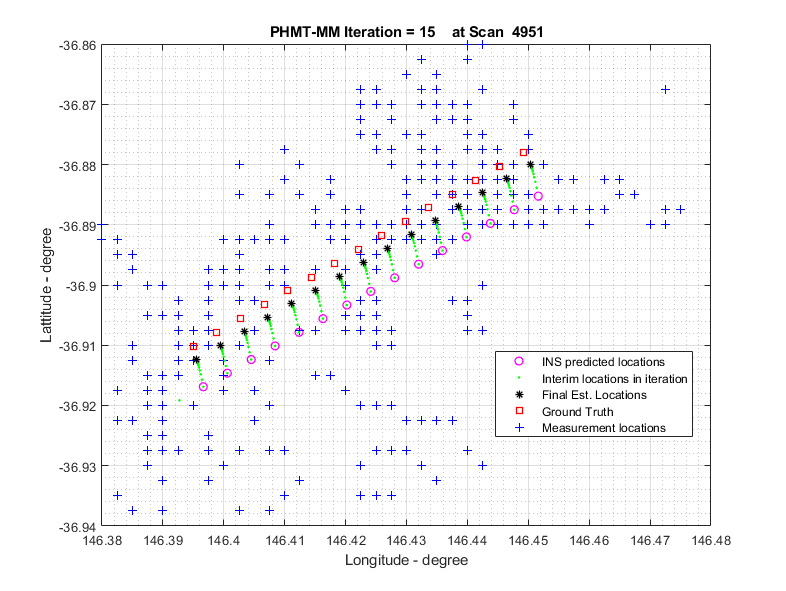}\\
  \caption{A snapshot of the PMHT-MM iteration process with gravity field map at the Scan 4951. The PMHT-MM starts the iteration from the INS predicted sensed locations (pink circles) and ends the iteration after 15 iterations at the final estimated locations (black stars) based on the the sequence of T=15 subsets of candidate measurement locations (blue plus symbols).}\label{fig-64}
\end{figure}

\subsubsection{Discussions}
\vskip 4 mm
\begin{itemize}
\item The experimental results show that a robust map matching performance is obtained by the PMHT-MM algorithm because it takes account of  both local and spatial data correlation to provide an estimate of  signal coordinates using  an EM approach. While it works in a batch mode, only relatively small amount of data samples are required to  obtain a reasonable signal location estimate. This is in contrast to  those map matching approaches, i.e., \citep{Wang2017,Wu2017} that require enormous measurements to carry out an area based cross correlation.

\item We plot a snapshot of the PMHT-MM iterative localisation process at a sampling epoch in Fig.~\ref{fig-64} in a run with aiding using the gravity field map. The plot shows the process that the algorithm drag the ``belief state (pink circle from INS)'' to the final estimated locations (black stars), i.e., to the state iteratively updated by the measurement location collection (blue plus symbols) with a linear kinematic constraints.

\item Gravimetric sensor noise levels have a direct impact on the accuracy and robustness of map matching aiding. The simulation results suggest that using a low noise field sensor, the PMHT-MM algorithm is able to use a batch state of small length to yield a robust aided inertial navigation performance without track divergence.
\item The proposed algorithm may be used for map matching with other type of sensor measurements, such as the gravity gradient tensor \citep{Jekeli2006}, magnetometer measurements \citep{Kim2019}, or terrain-based navigation \citep{Nygren2004},etc.
\end{itemize}

\section{Conclusions}
\label{sec5}

In this paper, we present a probabilistic multiple hypotheses tracking map matching algorithm for  gravimetric data map matching to aid an inertial navigation system in the absence of other aiding sources. The approach eliminates map measurement ambiguity by taking into account the kinematic constraints of the platform and permits incorporation of data maps with a range of  accuracy levels. Simulation results using online maps show the robustness and effectiveness of the algorithm for removing position drift that arises in INS over a long duration. Although the application shows an integrated gravimetric sensor map matching inertial navigation scenario, the algorithm is applicable to other map matching based applications with measurements under a low data sampling regime.

The proposed PMHT-MM solves the map matching localisation problem via an iterative batch processing procedure that handles map measurement ambiguity with kinematic constraints. As suggested by simulation results, it is capable of working at a low measurement rate with low resolution geophysical maps and giving a larger margin for trade-off between aiding robustness and ability of the tracker to handle vehicle maneuvers.

\section*{Acknowledgments}
The authors wish to acknowledge Professor Andrew Greentree of Quantum Physics from the school of science, RMIT university  for his very helpful comments and suggestions in this work.


\end{document}